%% file: main.tex
\documentclass[letterpaper, 10 pt, conference]{ieeeconf}
\IEEEoverridecommandlockouts                                       
\usepackage[utf8]{inputenc}
\usepackage{amsmath,amssymb,amsfonts}
\usepackage{bm}
\usepackage{url}
\usepackage{graphicx,graphics}
\usepackage{multirow}
\usepackage{booktabs}
\usepackage{multirow}
\usepackage[ruled,linesnumbered]{algorithm2e}
\usepackage{textcomp}
\usepackage{tabularx}
\usepackage{relsize}
\usepackage{soulutf8}
\usepackage[flushleft]{threeparttable}
\usepackage{cite}
\usepackage[table,dvipsnames]{xcolor} 
\usepackage{rotating}
\usepackage{xcolor}
\usepackage{comment}
\usepackage{soul}
\usepackage{tikz}
\input{macros.tex}

\newcommand*\circled[1]{\tikz[baseline=(char.base)]{
            \node[shape=circle,draw,inner sep=0.5pt] (char) {#1};}}
\newcommand{\myfigureshrinker}{\vspace{-0.4cm}}
\newcommand{\mytableshrinker}{\vspace{-0.3cm}}
\newcommand{\midsepremove}{\aboverulesep = 0.2mm \belowrulesep = 0.2mm}
    \midsepremove
\newcommand{\midsepdefault}{\aboverulesep = 0.605mm \belowrulesep = 0.984mm}
    \midsepdefault

\DeclareMathOperator*{\argmin}{arg\,min}
\graphicspath{{figures/}}
\DeclareGraphicsExtensions{.pdf,.jpeg,.png,.tiff,.jpg}
\title{\LARGE \bf Dual Quaternion-Based Visual Servoing \\for Grasping Moving Objects}%
\author{%
    Cristiana de Farias$^{1,\dagger}$, Maxime Adjigble$^{1}$, Brahim Tamadazte$^2$, Rustam Stolkin$^1$, Naresh Marturi$^1$% 
    \thanks{$^1$ Extreme Robotics Laboratory, School of Metallurgy and Materials, University of Birmingham, Edgbaston, B15 2TT, UK. $^{2}$ Sorbonne Universit\'e, CNRS UMR 7222, INSERM U1150, ISIR, F-75005, Paris, France. $^{\dagger}$Corresponding Author: \texttt{\{CXM1029\}@student.bham.ac.uk}}
    \thanks{This work was supported by the UK National Centre for Nuclear Robotics (NCNR), part-funded by EPSRC EP/R02572X/1 and in part supported by CHIST-ERA under Project EP/S032428/1 PeGRoGAM.}%
}%
\begin{document} \sloppy
\bstctlcite{IEEEexample:BSTcontrol}
\maketitle
%----------------- ABSTRACT ---------------
%
\input{sections/abstract.tex}
%
%----------------- INTRODUCTION ---------------
\section{Introduction}
\label{sec:intro}
\input{sections/introduction.tex}
\section{Problem Description}
\label{sec:problem}
\input{sections/problem.tex}
% %
%---------------- Overall METHOD ---------------
\section{Methodology}
\label{sec:method}
\input{sections/method.tex}
%%
%---------------Results-------------------------
\section{Experimental Results}
\label{sec:results}
\input{sections/results}
\section{Conclusion}
\label{sec:conclusion}
\input{sections/conclusion}
%
%-------------- Bibliography -------------------
\bibliographystyle{IEEEtran}
\bibliography{IEEEabrv,references}

\end{document}

%% file: macros.tex
 \def\dq#1{\underline{\boldsymbol{#1}}}

 \def\quat#1{\boldsymbol{#1}}

 \def\mymatrix#1{\boldsymbol{#1}}

 \def\myvec#1{\boldsymbol{#1}}

 \def\dual{\varepsilon}

 \def\norm#1{\left\Vert #1\right\Vert }

 \def\imi{\hat{\imath}}

 \def\imj{\hat{\jmath}}

 \def\imk{\hat{k}}

 \def\vector{\operatorname{vec}}

 \def\realset{\ensuremath{\mathbb{R}}}
 \def\SE#1{\ensuremath{SE(#1)}}
 \def\quatset{\ensuremath{\mathbb{H}}}
 \def\unitquatset{\mathcal{S}^{3}}
 \def\dualquatset{\mathbb{H}\otimes\mathbb{D}}
{} %
 \def\puredualquatset{\mathbb{H}_{0}\otimes\mathbb{D}}

 \def\unitdualquatset{\dq S}
{}

%% file: sections/abstract.tex
\begin{abstract}
This paper presents a new dual quaternion-based formulation for pose-based visual servoing. Extending our previous work on local contact moment (LoCoMo) based grasp planning, we demonstrate grasping of arbitrarily moving objects in 3D space. Instead of using the conventional axis-angle parameterization, dual quaternions allow designing the visual servoing task in a more compact manner and provide robustness to manipulator singularities. Given an object point cloud, LoCoMo generates a ranked list of grasp and pre-grasp poses, which are used as desired poses for visual servoing. Whenever the object moves (tracked by visual marker tracking), the desired pose updates automatically. For this, capitalising on the dual quaternion spatial distance error, we propose a dynamic grasp re-ranking metric to select the best feasible grasp for the moving object. This allows the robot to readily track and grasp arbitrarily moving objects. In addition, we also explore the robot null-space with our controller to avoid joint limits so as to achieve smooth trajectories while following moving objects. We evaluate the performance of the proposed visual servoing by conducting simulation experiments of grasping various objects using a 7-axis robot fitted with a 2-finger gripper. Obtained results demonstrate the efficiency of our proposed visual servoing. %
% This paper describes a new formulation of a position-based visual servoing (PBVS) using a dual quaternion algebra. Instead of using the traditional axis-angle parameterization usually used in the literature, dual quaternions allows designing the visual servoing task in a more compact manner, considering intuitively joint limits and singularities avoidance in a secondary null-space task. It is demonstrated that dual quaternion approach yields asymptotic regulation of the rotation and translation error as expected as well as a straight line trajectory of the camera.  
%
% The proposed methods are successfully validated in simulation by performing grasping tasks of moving objects. The obtained performances in terms of accuracy, behavior and stability are interesting to extend this approach by implementing the controller on an experimental station and a realistic scenario for robotic treatment of nuclear waste. 
\end{abstract}
%
% speaking on proof of stability in the abstract ? 
%In addition, a proof of stability is also provided.

%% file: sections/introduction.tex
% Over the years, robotic manipulation has been increasingly focusing on dynamic real-world applications, especially in unstructured environments.  %For this, robots need better and timely decision-making capabilities based on the real-time perception of the world. 
% For this, robots need to be integrated with controllers bearing superior decision-making abilities to cope with uncertainties in perceived information. 
% Visual sensors have been a popular choice of perception for tasks such as object tracking or grasping \cite{2008_VS_grasp1,1999_Chaumette_212VS,2018_Adjigble_LoCoMo}. Particularly, the integration of vision and manipulation is often fulfilled using visual servoing (VS) techniques \cite{1996_Hutchinson_VSTutorial,1999_Chaumette_212VS,2006_chaumette_visual_P1}. Indeed, this integration in real-time raises several challenges that often demands for an novel approaches and the use of state-of-the-art technologies. Nonetheless, existing techniques for visual servoing still explore different modeling, representation and control strategies for the vision and motion generation problems. In this work, we will take a completely different approach seeking to integrate the same state-of-the-art representation based on unit dual quaternion, that is often used for computer vision, to the visual servoing, manipulation control and grasping problem. 
Currently, many industries are increasingly using robotic manipulators on their production lines to perform pick and place of objects, e.g. bin picking, box packaging, etc. Over the years, grasping and manipulation of static objects has been well-explored, and a plethora of methodologies are presented in the literature~\cite{bicchi2000robotic, sahbani2012overview,bohg2013data, Farias_BayesianGrasping}. With the recent advancements in computing for industrial automation, few industries have adopted to use vision systems to localize, track and grasp planar-moving objects, e.g., from a moving conveyor. However, most of these systems still rely on vacuum suction and assume constant moving speed where uncertainties in perception (e.g., error in predicted object pose due to varied speed) or inaccurate robot modeling may lead to pick-up failures. Nevertheless, handling arbitrarily moving objects in a 3D space is an open challenge to solve mainly due to the difficulties associated with planning arm plus hand motions while simultaneously tracking the object trajectory. The methodologies designed to grasp static objects are not feasible and robots need to be integrated with controllers bearing superior decision-making abilities to cope with uncertainties in perceived exteroceptive information. 

\begin{figure}
    \centering
    \includegraphics[width=0.95\columnwidth]{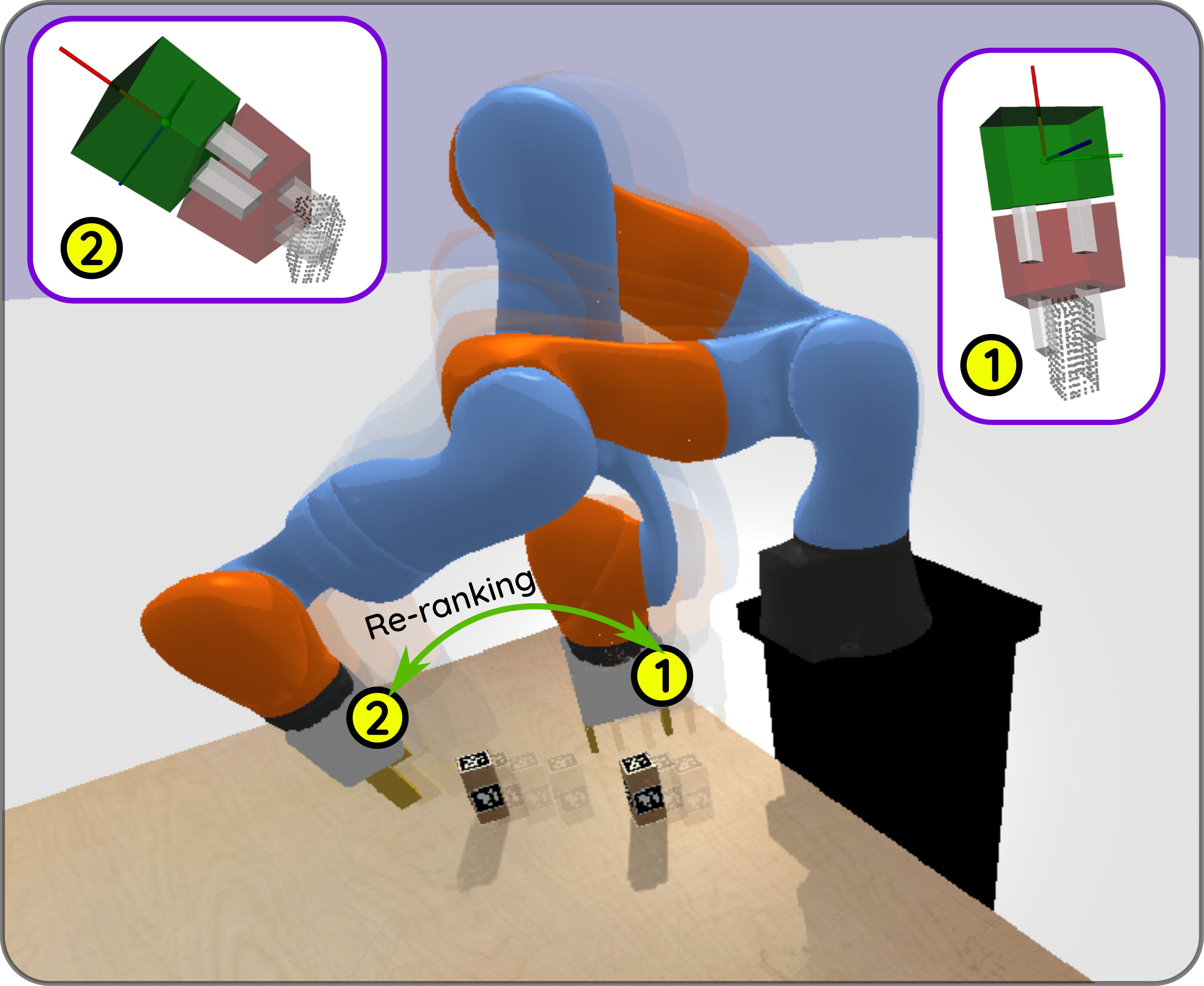}
    \caption{Grasping arbitrarily moving objects in 3D space using the proposed dual quaternion based visual servoing. Arrow represents grasp re-ranking between two poses marked with circled numbers. (inset) Screenshots showing grasp candidates generated by LoCoMo at these two poses. Green gripper represents pre-grasp and red gripper is the grasp. }
    \myfigureshrinker
    \label{fig:teaser}
\end{figure}

In this context, we present a singularity-robust dual quaternion-based visual servoing integrated with an efficient grasping algorithm to grasp arbitrarily moving objects in the robot’s task space as shown in Fig.~\ref{fig:teaser}. %
A variety of visual servoing methods have emerged in the literature, which are broadly classified into image-based (IBVS), position-based (PBVS) and hybrid approaches~\cite{2006_chaumette_visual_P1}. In this work, we are particularly interested in PBVS where the object’s relative position, estimated using perceived vision information, is used to control the manipulator’s movements. Multiple methods are available in the computer vision literature to estimate the full six Degrees of Freedom (DoF) pose (3 rotations and 3 translations) of an object in 3D space~\cite{marchand2015pose}. This is a vital step in any PBVS where the estimated object pose is used to regulate the error between current and desired positions of the robot’s end-effector. Consequently, any grasp planned on the object can be robustly updated or re-planned to the dynamic movements and perturbations. 

Most of the existing state-of-the-art in grasping are \textit{one-shot}, \textit{i.e.}, they perceive once and plan grasps assuming the object is stationary; the arm is then commanded to the grasp pose without using any feedback~\cite{bohg2013data}. However, they often fail in case of unplanned object movements or when there are uncertainties in the estimated hand to object relative poses. Few works have considered visual feedback to correct hand poses while reaching to grasp~\cite{marturi2019dynamic, gridseth2015visual, husain2014realtime, kim2014catching}. In~\cite{marturi2019dynamic}, Marturi et al. presented an approach to dynamically replan to grasp a moving object based on the vision information from two depth cameras, one hand mounted and the other scene camera. In~\cite{kim2014catching}, Kim et al. proposed a prediction-based programming by demonstration approach to catch objects in flight. An IBVS technique integrated with shared control grasping is presented in~\cite{gridseth2015visual}. Although, most of these methods have shown satisfactory performance in grasping moving objects, their efficiency depends on the availability of inverse kinematic solutions throughout the trajectory. In addition, especially in the presence of redundant joints, manipulators might get caught in singular configurations. 

Camera placement also plays a significant role in grasping moving objects. Two types of configurations are possible, \textit{i.e.}, \emph{eye-in-hand} where the camera is positioned on the robot end-effector; and \emph{eye-to-hand}, where the camera is placed somewhere in the scene observing both robot end-effector and task object. For dynamic grasping tasks, the latter is often preferred as it does not suffer ``blind spots’’ while visual tracking and is more biologically inspired. However, it still requires simultaneous tracking and pose estimation of both hand and object, which makes it challenging to use due to the lack of robust online tracking methodologies. As the main focus of this work is to design a new visual servoing controller, object pose estimation and visual tracking are not in its scope. For the sake of tracking object movements, we have used the conventional fiducial markers.%As this paper is more concerned with the design of a new visual servoing controller and not with pose estimation or tracking, we used fiducial markers for tracking and estimating object poses. 

We use dual quaternions as our main basis for the visual servoing in this paper. Similar to homogeneous transformation matrices  or angle-axis representations, dual quaternion can be used to represent full 6 DoF. Moreover, unit dual quaternions are singularity-free and have the benefit of being more compact and less computationally demanding than traditional representations, which makes them more suitable for visual servoing tasks. More specifically, they have proven to be an efficient and useful representation for many topics related to vision, such as hand-eye calibration \cite{1999_Daniilidis_HandEyeCalib}, simultaneous mapping and localisation~\cite{2018_LI_Slam} and pose estimation and tracking
~\cite{2020_Saltus_DQ_PVBS, 2005Wu_navigation,busam2017camera}. %
%
% Due to the demand for robust and cost-efficient applications in both computer vision and robotic, there has also been an increased interest in the study of kinematic representation and control based on dual quaternion algebra \textcolor{red}{[]}. Similar to homogeneous transformation matrices (HTM), unit dual quaternion are a singularity free, non-minimal representation for rigid body motions with coupled rotation and translation. Moreover, unit dual quaternion have the added benefit of being more compact and less computationally demanding than HTM \cite{1998Aspragathos_comparison } as well as providing a straightforward way to represent geometric primitives (such as Plüker lines or planes) which are useful to describing geometrical tasks directly on the task space \cite{2017adorno_DQbook}. More specifically, dual quaternion have proven to be an efficient and elegant and useful representation for many topics related to vision, such as hand-eye calibration \cite{1999_Daniilidis_HandEyeCalib}, simultaneous mapping and localisation \cite{2018_LI_Slam} and pose tracking and estimation \cite{2020_Saltus_DQ_PVBS, 2005Wu_navigation,busam2017camera} . 
%
Despite the increased interest on using dual quaternion for both robotic control and vision applications, literature which combines both is still scarce. %In \cite{2006Hu2_quaternion_VS}, the authors have presented a quaternion-based visual servoing schema for compensating rotational errors in case of calibration errors.
No known instances of using dual quaternion for visual servoing are reported in the literature until very recently a method has been presented by Saltus et al. in~\cite{2020_Saltus_DQ_PVBS}. %
%So far, only Saltus et al. in \cite{2020_Saltus_DQ_PVBS} have proposed a dual quaternion PVBS with pose tracking achieved using extended Kalman filter. 
There, the authors presented a PBVS schema to regulate the camera velocity with pose tracking achieved using an Extended Kalman filter. %

In this paper, we take a different approach for the design of a dual quaternion-based visual controller by explicitly integrating grasping into our visual servoing framework. With this aim, we explore our previous local contact moment (LoCoMo)-based unknown object grasping~\cite{2018_Adjigble_LoCoMo} to grasp arbitrarily moving objects in a 3D space. Firstly, LoCoMo will generate a ranked list of stable grasp poses on the perceived point cloud data of the scene object. Combining this with dual quaternion-based visual servoing, we will demonstrate the task of tracking and grasping arbitrarily moving objects. Capitalising on the dual quaternion spatial distance error between current end-effector pose and LoCoMo-provided pre-grasp pose, we built a dynamic grasp re-ranking metric to select the best feasible grasp for a moving object. In other words, the proposed visual servoing framework simultaneously analyses the quality of multiple grasp configurations and their distance to the gripper. Such analysis is performed in real-time, which allows the robot to switch between grasps as the object moves. Finally, as the robot dynamically tracks the object and switches among feasible grasps, it may encounter undesired configurations, such as singularities and joint limits. Therefore, the developed dual quaternion-based control strategy (while ensuring a singularity free representation) explores the robot null-space to provide a smooth motion avoiding joint limits. Overall, the main contributions of this paper are summarised as follows:
\begin{itemize}
    \item A dual quaternion-based PBVS scheme described on the manipulator's joint space, which is robust to singularities. This ensures smooth motion for the manipulator while tracking moving objects.
    \item We extend our previous work on stationary unknown object grasping based on local contact moments to grasp arbitrarily moving objects. For this purpose, we designed a new ranking metric based on dual quaternion error and vantage-point tree (vp-tree) to dynamically select the best feasible grasp during tracking trajectories.
\end{itemize}

The remainder of this paper is organised as follows. In Section~\ref{sec:problem}, we introduce the problem and present the developed method pipeline. Section~\ref{sec:method} presents the technical details of the developed method. Thereafter, in Section~\ref{sec:results}, we discuss the simulation experiments conducted using a 7-DoF robot fitted with a parallel-jaw gripper.

%% file: sections/problem.tex
\begin{figure*}
    \centering
    \includegraphics[width=0.95\textwidth]{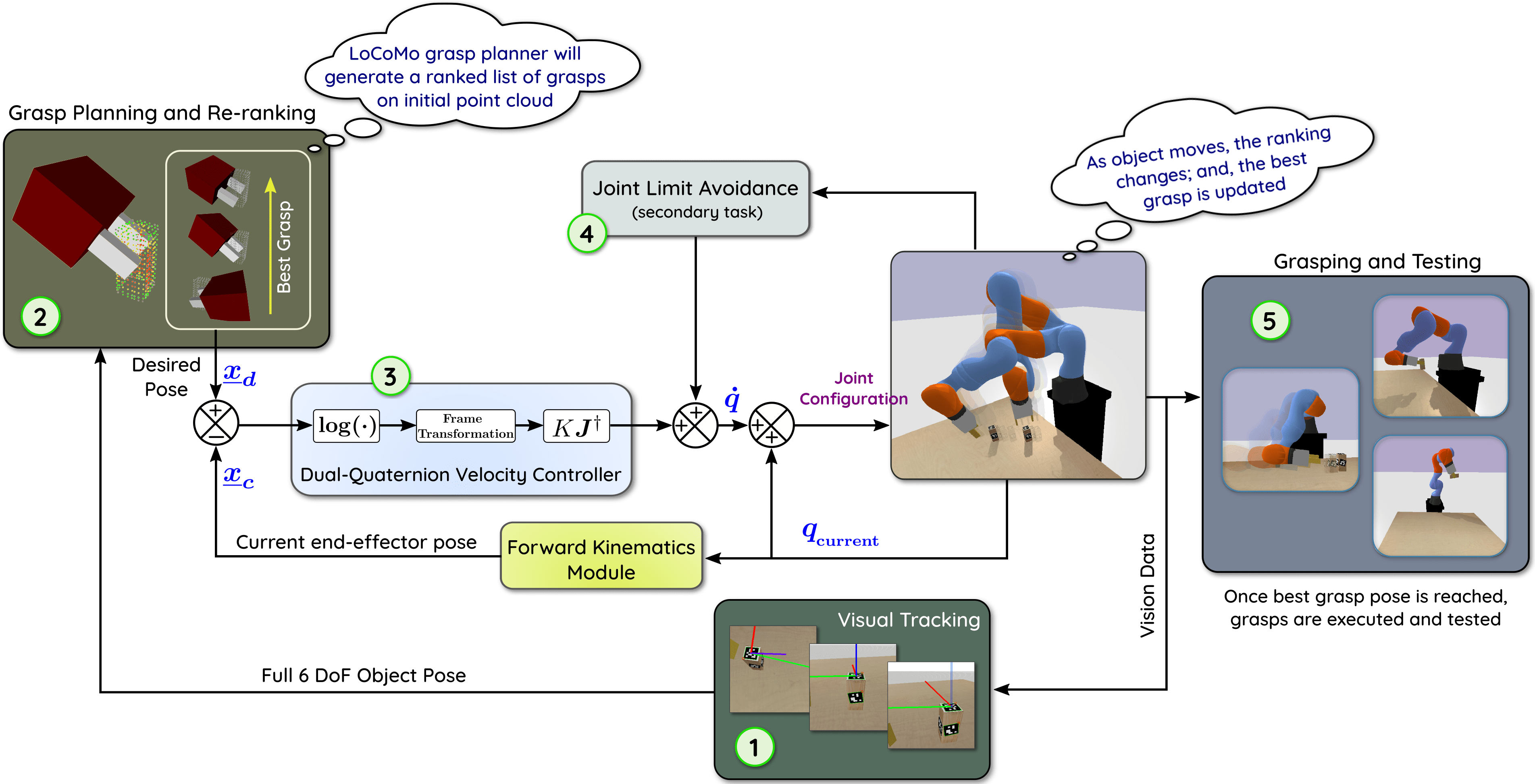}
    \caption{Control flow of the proposed method to grasp moving objects. It consists of five stages, which are marked with circled numbers. Using point cloud captured by a scene-mounted depth camera, LoCoMo grasp planner provides initial best grasp candidates of the object. Simultaneously, object pose is estimated by tracking fiducial markers placed on the object. Whenever the object moves, the desired pose for visual servoing, \textit{i.e.}, the pre-grasp pose computed by the grasp planner, is updated via re-ranking. Once the visual servoing converged and the object is grasped, the grasp stability is tested.}
    \label{fig:VS_flow}
    \myfigureshrinker
\end{figure*}
% ----
\subsection{Problem Statement}\label{subsec:Problem}
% -----
In general, PBVS problems aim at regulating the pose of a system based on visual information. Traditionally, these servoing systems are described using the axis-angle notation for rotations and have position and orientation decoupled. In this work, we tackle the problem of developing an efficient and generic robotic visual servoing strategy by designing a joint space visual regulator based on dual quaternion algebra. 

% Given an estimated object pose in dual quaternion notation $\dq{x}(\bm n(t))$, where $\bm n(t)$ is the vector of image features used to estimate the pose, and a reference pose $\dq x_d$, also depicted in dual quaternion notation, we can define the visual servoing problem as 
Following dual quaternion notations, let us assume that the current pose is $\dq{x}$ and the desired pose is $\dq x_d$, the visual servoing problem is defined as %, %where $\bm n(t)$  is the vector of image features used to estimate the pose. Using the reference or desired pose $\dq x_d$, the visual servoing problem is defined as
\begin{equation}
    \argmin\left [\dq e(\dq x, \dq x_d) \right] 
    \label{eq:error_generic}
\end{equation}
where, $\dq e$ is the dual quaternion error function for the pose. Hence, our problem becomes finding a control strategy which controls the manipulator velocities to minimise $\dq e$. %can be applied to a manipulator to minimise \eqref{eq:error_generic}. %
%
%Furthermore, in manipulation applications, one of the main goals is usually grasping an object. In order to address that, we add, as an extra element to our design, a dynamic grasp planner. 
Given the perceived point cloud of an object, LoCoMo provides feasible grasp ($\dq x_g$) and pre-grasp ($\dq x_{pg}$) poses. These are then used as visual servoing reference poses $\dq x_d=\dq x_{pg} | \dq x_{g}$. More details are presented in the next section. %In this manner, during visual servoing, the controller will first converge to the pre-grasp pose. After this, the reference pose is updated with the object grasp pose, \textit{i.e.}, $\dq x_d=\dq x_{g}$ so as to move the robot to execute grasps. 
The desired manipulator poses are automatically updated from the pose obtained via the object's visual tracking. Throughout this process, by using visual servoing instead of conventional path planning, we are able to accomplish moving object grasping.
%The desired manipulator poses are automatically updated with a change in the object pose that is obtained via visual tracking. Throughout this process, by using visual servoing instead of conventional path planning, we are able to accomplish moving object grasping.
%
\subsection{Method Overview}\label{subsec:method_overview}
Fig.~\ref{fig:VS_flow} depicts the pipeline of our dual quaternion-based visual servoing to grasp arbitrarily moving objects. It consists of five different stages (numbers inside circles in figure): \circled{$\bm 1$} Visual tracking from fiducial markers, \circled{$\bm 2$} grasp planning from LoCoMo~\cite{2018_Adjigble_LoCoMo}, \circled{$\bm 3$} velocity regulator and robot joint space visual servo control, \circled{$\bm 4$} joint limit avoidance as a secondary task projected at the robot null-space and \circled{$\bm 5$} grasping and testing. %
The process starts with \circled{$\bm 1$}, where we capture the object point cloud and an image to estimate initial object pose. It is worth noting that we use a scene mounted and \textit{hand-eye} calibrated RGB-D camera for this purpose. As mentioned earlier, we compute the pose by detecting the fiducial markers that are placed on the object's surface \cite{garrido2014automatic}. Alternatively, any other visual object pose tracker can be used. At each subsequent step, the object pose is updated and is passed to the grasp planning module. %used by our grasp re-ranking module, which selects the feasible grasp candidate,\textit{ i.e.}, the desired pose for visual servoing. 
In \circled{$\bm 2$}, for the first iteration, a list of best grasping and pre-grasping poses are generated by our LoCoMo grasp planner on the object point cloud. Pre-grasping poses are selected as the poses with a fixed offset on the approach axis, as shown in Fig.~\ref{fig:teaser}. For the subsequent iterations, as the object moves, the best grasping pose may change due to kinematic constraints of the robot. Therefore, given the current manipulator configuration, at each step, the grasping poses are dynamically re-ranked and the best pre-grasping pose is used by the visual servoing as the desired pose $\dq x_d$ to reach. Once the robot is at pre-grasping pose, the grasping pose is used as desired pose such that the robot can proceed to grasp the object. This two-point reach reduces the chance of collision between the end-effector and the object on its approach. %. The best (pre)grasping pose is then sent to the dual quaternion-based visual servoing control as the reference pose $\dq x_d$. 
In \circled{$\bm 3$}, we have the main controller that minimises the dual quaternion error as in \eqref{eq:error_generic}. The error here is defined as the dual quaternion spatial distance between the end-effector's current pose and the desired pose from previous step. %relative pose from the camera (or from the end-effector) to the object. 
Additionally, to avoid joint limits, we add a secondary task projected to the robot's null space. This is depicted by \circled{$\bm 4$} in Fig. \ref{fig:VS_flow}. Finally, in~\circled{$\bm 5$}, we execute the grasp and test its stability by performing a series of tests as described in~\cite{BenchmarkERL}. %lift the object, as well as test the stability of the grasp.%
All the technical details regarding our dual quaternion visual controller, grasp generation by LoCoMo and our re-ranking schema are described in the next sections.
%We further explain the control equations from steps \circled{$\bm 3$} and \circled{$\bm 4$} in Section \ref{sec:method}. The grasp planning from $\bm 2$, re-ranking procedure and its integration to the controller will be explained in Section \ref{sec:grasping}. And in Section \ref{sec:results} we show the validation of our method for reaching and grasping dynamical objects.
% \subsection{Notation}
% %
% In this paper, we follow the notation similar to that of in \cite{vilhenaadorno:hal-01478225}. All these are summarised below: 

%% file: sections/method.tex
This section provides compact formulations of our PBVS schema using dual quaternion algebra instead of the traditional axis-angle notation~\cite{2006_chaumette_visual_P1}. %Exploiting dual quaternion algebra and robot's null space, our VS design efficiently handles singular configurations producing smoother trajectories.
%
% --------------
\subsection{Mathematical Preliminaries}\label{sec:mathPreliminaries}
% --------------
%
Quaternions are an extension of imaginary numbers, often used to define three-dimensional rotations. First, let us consider $\imi,\imj,\imk$ be the three quaternion unit vectors such that $\imi^{2}=\imj^{2}=\imk^{2}=\imi\imj\imk=-1$. The algebra of quaternions~\cite{hamilton1866elements} is generated by the basis elements $1$, $\imi$, $\imj$, and $\imk$, yielding the non-commutative set%
\begin{equation}
    \quatset\triangleq\left\{ \eta+\myvec{\mu}:\ \myvec{\mu}{=}\mu_{1}\imi+\mu_{2}\imj+\mu_{3}\imk,\ \eta,\mu_{1},\mu_{2},\mu_{3}{\in}\realset\right\} \label{eq:quatset}
\end{equation}
where, for any quaternion $\quat{x}\in \quatset$, $\eta$ and $\myvec{\mu}$ are respectively the real and imaginary (or vector) parts. The quaternion conjugate is then defined as $\quat x{}^{\ast}\triangleq \eta-\myvec{\mu}$, which in turn allows to define the quaternion norm as $\norm{\quat x}^2={\quat x\quat x^{\ast}}={\quat x^{\ast}\quat x}$. Particularly, we define the subset of unit norm quaternions $\unitquatset\triangleq\left\{ \quat x\in\quatset: \,\norm{\quat x}=1\right\}$ under multiplication to represent spatial rotations, and the pure quaternions subset $\quatset_{0}\triangleq\left\{ \quat x_{0}:\eta+\myvec{\mu}\in\quatset\,,\,\eta=0\right\} $, to represent spatial translations. Those two subsets have important properties in representing rigid body motions. The vector isomorphism under $\quatset_0$ allows for operations such as inner and vector products and the group multiplication of unit quaternions. This guarantees that, for any $\quat{x}_1, \quat{x}_2\in\unitquatset$, the quaternions multiplication $\quat{x}_3 = \quat{x}_1\quat{x}_2 \in \unitquatset$. % will be such that $\quat{x}_3 \in \unitquatset$.%
%Another important aspect of unit quaternions is its geometric interpretation. In this case, 
Besides, a unit quaternion with rotation $\phi$ around the axis $\quat n\in \quatset_0$, can be given by
\begin{equation}
\quat x = \text{cos}\left(\frac{\phi}{2}\right)+\quat n\text{sin}\left(\frac{\phi}{2}\right) \label{eq:rotation_quat}
\end{equation}

Let $\quat x_1, \quat x_2 \in \quatset_{0}$. The quaternion inner and vector products are%
\begin{eqnarray}
\nonumber
    \langle \quat{x}_1, \quat{x}_2\rangle & \triangleq & -\frac{ \quat x_1\quat x_2+\quat x_2\quat x_1}{2}\\
    \quat {x}_1 \times \quat{x}_2 & \triangleq &  \frac{ \quat x_1\quat x_2-\quat x_2\quat x_1}{2}
    \label{eq:inner_product}
\end{eqnarray}

A further extension of imaginary numbers is the dual quaternion, which is introduced as non-minimal, non-commutative, singularity-free description of rigid body motions~\cite{selig2004geometric}. Dual quaternions are defined as the set 
\begin{equation}
    \dualquatset\triangleq\left\{ \dq x=\quat{x_{P}}+\varepsilon\quat{x_{D}}\ |\ \quat{x_{P}},\quat{x_{D}}{\in}\quatset\right\} \label{eq:dq_set}
\end{equation}
where, $\dual$ is called dual unit with $\dual^{2}=0,\,\dual\neq0$, and  $\quat{x_P}$ and $\quat{x_D}$ are the primary and dual parts of the dual quaternion $\dq x$, respectively. Furthermore, analogous to quaternions, the dual quaternion conjugate is $\dq x^{\ast}\triangleq\quat x_P^{\ast}+\dual\quat x_D^{\ast}$ and subsequently, $\norm{\dq x}^2={\dq x\dq x^{\ast}}$. Next, the unit dual quaternion subset under multiplication, $\unitdualquatset\triangleq\left\{\dq x\in\dualquatset:\,\norm{\dq x}=1\right\}$, forms the Lie group $\mathrm{Spin}(3)\ltimes\mathbb{R}^{3}$ that double covers $\SE{3}$. % and whose identity element is 1. 
The group inverse of $\dq x\in\unitdualquatset$ is $\dq x^{\ast}$. Finally, we define the pure dual quaternion as $\puredualquatset\triangleq\left\{ \dq x_{0}=\quat{x_{P}}+\varepsilon\quat{x_{D}}\ |\ \quat{x_{P}},\quat{x_{D}}{\in}\quatset_{0}\right\} $ such that for $\dq x_0\in\puredualquatset$, $\dq x_0^\ast=-\dq x_0 $.

Let us consider $\quat{x}_t\in \quatset_0$ and $\quat{x}_r \in \unitquatset$ being an arbitrary translation and rotation of a rigid body. The full rigid body displacement can then be described as follows
\begin{equation}
\dq x = \quat{x}_r+\dual\frac{1}{2}\quat{x}_t\quat{x}_r, \label{eq:rigid motion}
\end{equation} 
and the kinematic equation for the body twist will be
\begin{equation}
\dq {\dot x} = \frac{1}{2} \dq x\ \dq{\omega}^B, \label{eq:twist}
\end{equation} 
where, $\dq{\omega}^B\in\puredualquatset$ is the body twist and $\dq {\dot x}$ is the time derivative of $\dq x$. When dealing with rigid body motions, it is important to efficiently describe transformations between different frames. Let %$\dq{\omega}_b$ be the representation of the twist in the body frame and
$\ {}^I\dq{x}_B\in \unitdualquatset$ be the transformation between the inertial frame $\mathbb{F}^I$ and the moving body frame $\mathbb{F}^B$. Then inertial twist is given by,
\begin{equation}
    \dq{\omega}_I={}^I\dq{x}_B\dq{\omega}^B{}^I\dq{x}_B^\ast
    \label{eq:intertial_twist}
\end{equation}
%
%\eqref{eq:intertial_twist} represents the adjoint transformation, which is defined as

The adjoint transformation is given by
\begin{equation}
    \text{Ad}(\dq{x})\dq{y} \triangleq \dq{x}\dq{y}\dq{x}^{\ast}, \label{eq:adjoint}
\end{equation}
with $\dq{x}\in \unitdualquatset$ and $\dq{y}\in\puredualquatset$. Using \eqref{eq:adjoint}, the inertial twist can be re-written as $\dq{\omega}_I=\text{Ad}(^I\dq{x}_B)\dq{\omega}^B$, or in \eqref{eq:twist}, as $\dq{\dot x} = \frac{1}{2} \ \dq{\omega}^I\dq x$. Additionally, the logarithm operation involved within unit dual quaternion is
\begin{equation}
    \text{log}(\dq x) = \frac{\phi\quat n }{2}+\dual\frac{\quat p}{2}, \label{eq:DQ_log}
\end{equation}
%
%with $\quat n$ and $\phi$ being the rotation axis and angle of the unit dual quaternion and $\quat 
where, $\quat p$ represents translation. Finally, we define an operator for one-by-one mapping $\vector_6:\puredualquatset\rightarrow\realset^{6}$ and its inverse operation $\underline{\vector}_6:\realset^{6}\rightarrow\puredualquatset$.
%
% --------------
\subsection{Controller Design} \label{sec:ControllerDesing}
% --------------
%
%Now that the dual quaternion representation is defined, it is possible to express the new visual servoing controller in order to control the robot arm in its joint space. 
Now, we show how dual quaternions are used to express the visual servoing controller. Let us consider $\dq {x}_c$ to be the current pose of the end-effector, obtained from the robot's forward kinematic model. Following \eqref{eq:error_generic}, the error $\dq{e}$ to be regulated is given by
%
% It is possible to regulate the error in \eqref{eq:error_generic} to zero by the visual servoing controller define the dual quaternion spatial error used in function \eqref{eq:error_generic} to be regulated to zero by the visual servoing controller. It is given by the following 
%Let us consider $\dq x_o$ be the dual quaternion representation of the object pose, estimated from the fiducial markers, as explained in Section~\ref{subsec:method_overview}. Also, let $\dq {x}_c$ be the current estimated pose of the end-effector, obtained from the robot's forward kinematic model. Then the relative pose from the gripper to the object will be $\dq {x}^{o}_c=\dq{x}^\ast_c\dq{x}_o$. Thus, it is possible to define the quaternions error function of \eqref{eq:error_generic}  (\textcolor{red}{where is located this equation (1)?} to be regulated to zero by the visual servoing controller. It is given by the following 
\begin{equation}
    \dq{e} = \left (\dq{x}_c\right )^\ast\dq x_d. \label{eq:error}
\end{equation}
where, $\dq x_d$ is the reference pose. Selection of the reference pose is explained in the next section. It is worth noting that all operations are performed with respect to the robot base coordinate frame. For this purpose, the visually estimated poses (grasps) are transformed to the base frame using camera-robot calibration. For more details, we refer the readers to~\cite{Chaumette07b}.

In order to minimise the error and ensure convergence, we define a logarithmic controller that leads to a closed loop system, such as $\dq {\dot{e}} = -\frac{1}{2} \text{log}\left(\dq e \right)\dq e$. The stability of such a system is discussed in~\cite{han2008_quaternionControl}. Thereby, using \eqref{eq:twist},
%
%In this manner, we may directly compute the pose's first order derivative as $\dq {\dot x}_c$. Thereby, from \eqref{eq:twist}, it is possible to derive current pose as follows
%In order to minimise the error and ensure convergence, we define a logarithmic controller that leads to a closed loop system, such as $\dq {\dot{e}} = -\frac{1}{2} \text{log}\left((\dq e) \right)\dq e$. This system has been proven stable in~\cite{han2008_quaternionControl}. In this manner, we may directly compute the pose's first order derivative as $\dq {\dot x}^{o}_c\approx \dq{\dot x}^\ast_c\dq{x}_o $, with the assumption that $\dq x_o$ constant. Thereby, from \eqref{eq:twist}, it is possible to derive camera pose as follows
%
\begin{equation}
    \dq {\dot x}_c =  \frac{1}{2} \dq x_c\ \dq{\omega}^B_c   \label{eq:rel_kin_eq}
\end{equation}
%\begin{equation}
%     \dq {\dot x}c\approx \left( \frac{1}{2} \dq x_c\ \dq{\omega}^B_c \right)^\ast\dq{x}_o = -\frac{1}{2}\dq{\omega}^B_c\  \dq x_c^\ast  \dq{x}_o=  -\frac{1}{2}\dq{\omega}^B_c\  \dq {x}^{o}_c, \label{eq:rel_kin_eq}
% \end{equation}
in which $\dq \omega^B_c$ is the current end-effector twist. Hence, from \eqref{eq:error}, the error's first order time derivative is $\dq {\dot e} = \left (\dq{\dot x}_c\right )^\ast \dq x_d $. Now, substituting for $\dq {\dot x}_c$ using \eqref{eq:rel_kin_eq}, we get
\begin{equation}
    \dq {\dot e} = \left (\frac{1}{2} \dq x_c\ \dq{\omega}^B_c \right )^\ast \dq x_d = - \frac{1}{2}\  \dq{\omega}^B_c \dq {x}_c^\ast  \dq x_d = - \frac{1}{2}\  \dq{\omega}^B_c \dq e
    \label{eq:error2}
\end{equation}
Finally, given the desired closed loop dynamics, the current twist in the end-effector body frame can be defined as 
\begin{equation}
  \begin{gathered}
     -\frac{1}{2} \text{log}\left(\dq e \right)\dq e = -\frac{1}{2}\  \dq{\omega}^B_c \dq e \\
     \dq{\omega}^B_c = \text{log}(\dq e)
     \label{eq:contro_input}
  \end{gathered}
\end{equation}

%To perform visual servoing with a robotic manipulator we need to apply the control input in \eqref{eq:contro_input} to the robot joint space. Particularly to our design, we choose to control the robot's global pose, thus we must work with the input twist in the world frame, that is $\dq{\omega}^I_c=\text{Ad}(\dq x_c)\dq{\omega}^B_c$. Furthermore, in most robotic systems, the camera frame will also not coincide with the end-effector frame and in order to add precision to our system we must account for this extra transformation. Thus, let $\dq {x}_{c|\text{EF}}$ be the transformation from the camera to end-effector frame, then the velocity will be $\dq \omega_{\text{EF}}=\dq {x}_{c|\text{EF}}\dq{\omega}^I_c$. In \ref{fig:VS_flow} these operations are represented by the "frame transformation block". To ensure simplicity, from now the inertial twist will be simply denoted as $\dq \omega$  and unless stated, all variables will be given in the world frame.

To perform visual servoing with a robotic manipulator we need to apply the control input in~\eqref{eq:contro_input} to the robot joint space. With our design, we choose to control the robot's global pose; thus, we work with the input twist in the robot base frame, \textit{i.e.}, using~\eqref{eq:adjoint}, $\dq{\omega}^I_c=\text{Ad}(\dq x_c)\dq{\omega}^B_c$. %Furthermore, in most robotic systems, the camera frame will also not coincide with the end-effector frame and in order to add precision to our system we must account for this extra transformation. Thus, let $\dq {x}_{c|\text{EF}}$ be the transformation from the camera to end-effector frame, then the velocity will be $\dq \omega_{\text{EF}}=\dq {x}_{c|\text{EF}}\dq{\omega}^I_c$.
In Fig.~\ref{fig:VS_flow}, this operation is represented by the "frame transformation block" in stage \circled{$\bm{3}$}. For the sake of simplicity, from now on, inertial twist will be simply denoted as $\dq \omega$  and unless stated, all variables are in the robot base frame. %Now, $\mymatrix J\in\realset^{6\times N_{\text{DoF}}}$ is the dual quaternion-based geometric Jacobian, which maps the coupled dual quaternion twist to the robot joint space as in \textcolor{red}{[reference]}. Then, the joint space controller can be defined as 
Considering $\mymatrix J\in\realset^{6\times N_{\text{DoF}}}$ the dual quaternion-based geometric Jacobian, which maps the coupled dual quaternion twist to the robot joint space as in~\cite{2018Figueredo}, the joint space controller is defined as
\begin{equation}
    \begin{gathered}
    \myvec{q}_{k+1} = \myvec{q}_k + K\mymatrix{J}^\dagger\vector_6\left( \dq \omega \right )\\
    \myvec{q}_{k+1} = \myvec{q}_k + K\mymatrix{J}^\dagger\vector_6\left( \text{Ad}\left( \dq x_c  \right )\text{log}(\dq e) \right ) \label{eq:jointController}
    \end{gathered}
\end{equation}
% \begin{equation}
%     \myvec{q}_{k+1} = \myvec{q}_k + K\mymatrix{J}^\dagger\vector_6\left( \dq \omega \right ) \nonumber
% \end{equation}
% %
% \begin{equation}
%     \myvec{q}_{k+1} = \myvec{q}_k + K\mymatrix{J}^\dagger\vector_6\left( \text{Ad}\left( \dq x_c  \right )\text{log}(\dq e) \right ) \label{eq:jointController}
% \end{equation}
where, $\myvec{q}_{k+1}$ and $\myvec{q}_k$ are the commanded and the current joint configurations, respectively, $K>0$ is the proportional gain, and $\mymatrix J^\dagger$ is the Jacobian damped pseudo-inverse computed as $\mymatrix J^\dagger=\mymatrix J^T\left (\mymatrix J \mymatrix J^T + \lambda\mymatrix I_6 \right )^{-1}$, with $\lambda$ being a small scalar gain. 
%Note that if the robot is not redundant, $J^{-1}$ is sufficient. 
%with $\myvec{q}$ being the robot joint vector at the moment $k$, $k+1$ and $K$ the controller's proportional gain. 
%Moreover, $\mymatrix J^\dagger$ is the Jacobian damped pseudo-inverse, defined as $\mymatrix J^\dagger=\mymatrix J^T\left (\mymatrix J \mymatrix J^T + \lambda\mymatrix I_6 \right )^{-1}$, with $\lambda$ being a small scalar gain. We choose this inversion method for the Jacobian in order to add extra robustness to the manipulation strategy, ensuring the avoidance of regions with kinematic singularities.

%As our aim is to design a controller that is robust to singularities, it is interesting to explore the manipulator's null-space for the implementation secondary tasks, such as avoiding joint limits. 
Since we consider a redundant robot in this work, we explore manipulator’s null-space to avoid joint limits. For this purpose, we define a secondary task joint-space mapping as $\mymatrix J_{s} = \myvec{q} - \myvec{\Bar{q}}$, with $\myvec{\Bar{q}}$ being the mean position for each joint. The secondary task cost to be minimized can be defined as $\mathcal{C}_{\mathrm{null}} = \frac{1}{2}\sum{\left ( \myvec{q} - \myvec{\Bar{q}} \right )^2}$. Using the Jacobian $\mymatrix J$, we define a null-space projector as $\mymatrix P = \left ( \mymatrix I_7 - \mymatrix J^\dagger \mymatrix{J} \right )$. Now,  \eqref{eq:jointController} with the added null-space term becomes 
\begin{equation}
     \myvec{q}_{k+1} = \myvec{q}_k + K\mymatrix{J}^\dagger\vector_6\left( \dq\omega \right )+K_s\mymatrix P \mymatrix{J}^{-1}_s \mathcal{C}_{\mathrm{null}}, \label{eq:nullspace_control}
\end{equation}
with $K_s < 0$ being the secondary task gain. Note that the secondary task will not have any effect on the primary visual servoing task.
%
%---------------- Grasping ---------------
\subsection{Grasp Planning and Re-ranking}
\label{sec:grasping}
\input{sections/grasping.tex}

%% file: sections/grasping.tex
Moving object grasping is performed by combining a grasp generator with the proposed visual servoing control scheme. As mentioned earlier, we have used LoCoMo-based grasp planner~\cite{2018_Adjigble_LoCoMo} to synthesise grasps on a task object. %
%Various grasp generation strategies have been proposed in the literature \cite{bicchi2000robotic, bohg2013data, shimoga1996robot, sahbani2012overview}. Our method make use of the Local Contact Moment (LoCoMo) metric presented in \cite{2018_Adjigble_LoCoMo} to generate a set of grasp poses for a given object.  
The advantage of using the LoCoMo grasp planner is two-fold. Firstly, the method neither requires any object 3D models nor offline training, making it easy to integrate with our visual servoing. Secondly, the grasp ranking scheme can be extended to provide dynamic grasp selection~\cite{adjigble2019assisted}, which is ideal for moving object grasping. %Our method uses a simulated depth camera to acquire the point cloud of the object. %
Given an object point cloud, an initial set of grasp and pre-grasp poses are generated using the LoCoMo grasp planner. The top-ranked grasp is selected and the visual servoing controller is initialized using this pose as the reference $\dq x_d$. Grasp computation and ranking is given by
\begin{equation}
\label{eq:locomo}
\begin{aligned}
        \mathcal{C}_{i} &= \frac{1}{N_{s}} \sum_{j=1}^{n}\left(\left(2\pi\right)^{n}\left |\Sigma  \right |\right) ^{\frac{1}{2}}\mathcal{N}(\Psi_j; \vec{0},\Sigma ) \\ 
        \mathcal{R} &= \gamma\prod_{i=1}^{N_{f}}\mathcal{C}_{i}^{\omega_{i}}
\end{aligned}
\end{equation}
where, $\mathcal{C}_{i}$ is the contact moment of each gripper finger $i$, $\mathcal{R}$ is the ranking score, $N_s$ is a normalizing term, $n$ is the number of point cloud patches sampled around the current location of the finger, $\mathcal{N}(\Psi_j; \vec{0},\Sigma )$ is the multivariate Gaussian density function centered at $\vec{0}$ with co-variance $\Sigma$, $\Psi_j$ is the difference of zero-moment shift vectors between object and gripper fingers surface, $\omega_i$ and $\gamma$ are weighting factors, and $N_{f}$ is the total number of fingers. More details can be found in~\cite{2018_Adjigble_LoCoMo}.

While the object is moving, we visually track its pose, given by $\dq x_o$, with a scene mounted camera. This in turn is used to update feasible grasps accordingly. To enable our controller to be as efficient as possible for dynamic grasping, a grasp re-ranking scheme is implemented. At each iteration, we select the $\mathcal{K}$ feasible grasps to the current gripper position using the dual quaternion distance. By feasible, we mean that there is an inverse kinematics solution from the current robot configuration. These set of grasps are then re-ranked and the top ranked grasp is used for visual servoing. 
% Given an object point cloud, grasps are generated and ranked using the metric $ \mathcal{R}$ computed by \eqref{eq:locomo}.
% %
% \begin{equation}
% \label{eq:locomo}
% \begin{aligned}
%         \mathcal{C}_{i} &= \frac{1}{N_{s}} \sum_{j=1}^{n}\left(\left(2\pi\right)^{n}\left |\Sigma  \right |\right) ^{\frac{1}{2}}\mathcal{N}(\varepsilon_j; \vec{0},\Sigma ) \\ 
%         \mathcal{R} &= k\prod_{i=1}^{N_{f}}\mathcal{C}_{i}^{\omega_{i}}
% \end{aligned}
% \end{equation}
% %

Even though all the generated grasps could be used for the re-ranking stage, using the $\mathcal{K}$ closest grasps allow to devise a more efficient algorithm, especially for real-time applications. We use a vantage-point tree (vp-tree)~\cite{yianilos1993data} for searching the nearest grasps to the robot's gripper. The vp-tree provides a fast and efficient nearest neighbor algorithm that can be used with any arbitrary metric space. For our method, we use the norm of the dual quaternion spatial error between two dual quaternions, e.g., for two poses $\dq x_1, \dq x_2$, this is computed as $\dq x_{e}=1-\dq x^{\ast}_{1} \dq x_{2}$ (then mapped to $\mathbb{R}^8$ where we take the Euclidean norm) in conjunction with the vp-tree to find the closest grasps. Using the $\mathcal{K}$ closest grasps and dual quaternion distances $dq_i$ of the grasp pose $i$ from the current gripper position, the grasp re-ranking $\Upsilon$ is defined as
\begin{equation}
        \Upsilon = \frac{dq_i}{dq_{max} - dq_{min}} \:\text{, with  ~} dq_{max} \neq dq_{min}
 \label{eq:grasp_reranking}
\end{equation}
where, $dq_{min}$ and $dq_{max}$ are respectively the minimum and maximum dual
quaternion distances between grasps and gripper pose among the $\mathcal{K}$ closest grasp candidates. $\Upsilon$ is updated at each iteration  and the highest ranked grasp is used for the visual servoing controller. This technique, however, can create chattering problem in the controller if the best grasp changes too frequently due to noise. To account for that we have set a hysteresis to the switching strategy, \textit{i.e.}, we only change the control reference pose if the new grasp ranking has improved at least by a switching threshold $\delta$~\cite{marturi2019dynamic}. This strategy allows to have a smooth and dynamic reach-to-grasp trajectory by using the closest best grasp candidate available.

%% file: sections/results.tex
In this section we present the simulation experiments conducted to validate the performance of our proposed dual quaternion based visual servoing and its ability in grasping arbitrarily moving objects. 
\subsection{Simulation Setup}
All experiments are performed using the open source Python wrapper for Bullet physics engine, Pybullet, which has been proved to be an efficient and stable simulation tool for robotics \cite{coumans2016pybullet}. Our robotic setup consists of a 7-DoF robot arm fitted with a parallel jaw gripper. For experiments we have selected three different objects from the YCB object dataset \cite{ycb} that are both suitable with our gripper and allow for marker placement on their surfaces. In order to make the visual servoing system more realistic, we have also simulated a virtual  rgb-d camera for both pose tracking and point cloud acquisition. Initially, we move the camera to three different locations around the object and stitch the clouds captured at these locations to obtain a task point cloud for LoCoMo. Besides, for the sake of visual object pose tracking, we have used ArUco fiducial marker tracking \cite{garrido2014automatic} with markers placed on each object. Object CAD models downloaded from the YCB dataset with altered textures to include the makers have been used for the simulation. We report the results following the evaluation protocol presented in \cite{BenchmarkERL}. That is, we simulate different trajectories for each object to move and attempt to dynamically grasp them. Once grasped, we perform various tests as in \cite{BenchmarkERL} to evaluate grasp stability.
\subsection{Controller Convergence Analysis}
\begin{figure}
    \centering
    \includegraphics[width=\columnwidth]{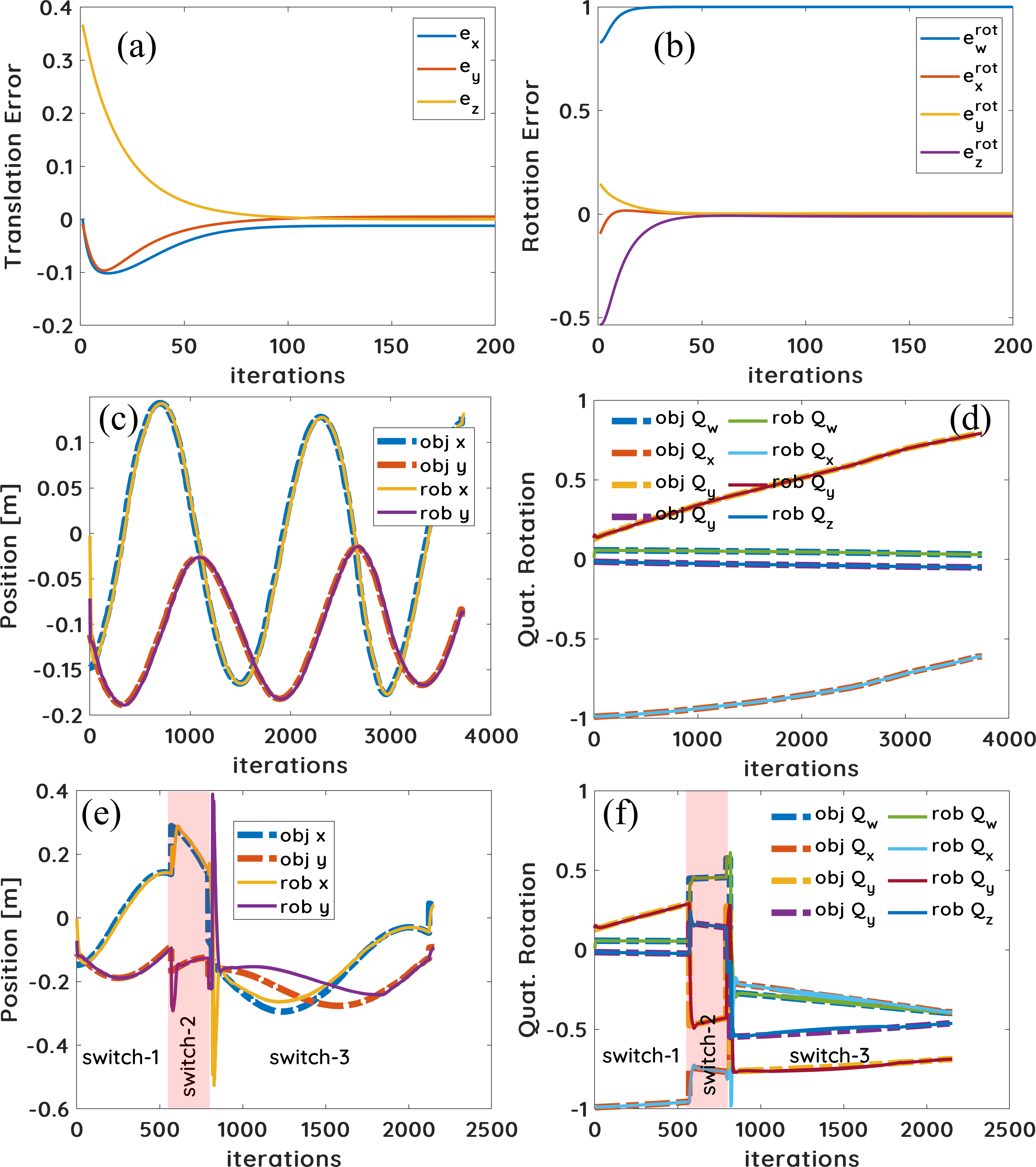}
    \caption{Plots describing the convergence of the proposed method. Top row shows the (a) translation and (b) rotation errors convergence when approaching the desired grasp configuration with respect to a stationary object. In the second row, we show the robot tracking an object moving in a circular trajectory with the grasp re-ranking disabled; (c) and (d) shows respectively position and orientation errors between the end-effector and desired pre-grasp pose. In the last row, we show (e) the position and (f) orientation errors with the proposed method when the grasp re-ranking is enabled. The displayed trajectory contains three grasp switches, \textit{i.e.,} re-ranking module adjusted the feasible grasp pose three times.}
    \label{fig:Convergence}
    \myfigureshrinker
\end{figure}
With this set of experiments, performed using ``woodblock’’ object (YCB ID: 45), we test the convergence of our visual servoing in case of both static and moving objects. The plots shown in Fig. \ref{fig:Convergence}(a) and \ref{fig:Convergence}(b) respectively depict the translational and rotational evolution of our method for a stationary object. These results clearly show smooth convergence. %Nevertheless, the rate of convergence can be improved by optimising the controller as in \cite{Chaumette07b}. %
 Next tests are done to validate and demonstrate the convergence while following moving objects. Here, we test the behaviour in case of both grasp re-ranking enabled and disabled. When the re-ranking is disabled, the system might run close to singularities and joint limits, and we analyse the performance of our controller in such harsh conditions. Plots in Fig.  \ref{fig:Convergence}(c) and \ref{fig:Convergence}(d) show how our velocity regulator given by \eqref{eq:nullspace_control} can simultaneously follow same grasp without re-ranking. Comprehensively, it performed well where robot closely followed the trajectory. Plots in Fig.  \ref{fig:Convergence}(e) and \ref{fig:Convergence}(f) show the behaviour when re-ranking given by \eqref{eq:grasp_reranking} is enabled. Particularly, we show the desired configuration switching to three different grasps, with the last one remaining stable until the moment of grasping (at the end of plot). These results clearly demonstrate the efficiency of our controller. In case of grasp switching, the controller successfully stabilised mid-trajectories and continued tracking the object. Particularly, we note that the orientation error in both Fig. \ref{fig:Convergence}(d) and Fig. \ref{fig:Convergence}(f) remain low. That is one of the advantages of using the dual quaternion representation, which has coupled rotation and translation. Overall, the quick stabilisation to perturbations and robustness to singularities and joint limits make our dual quaternion-based visual servoing well suited for the problem of grasping moving objects.%
\subsection{Grasping Arbitrarily Moving Objects}
\begin{figure*}
    \centering
    \includegraphics[width=0.8\textwidth]{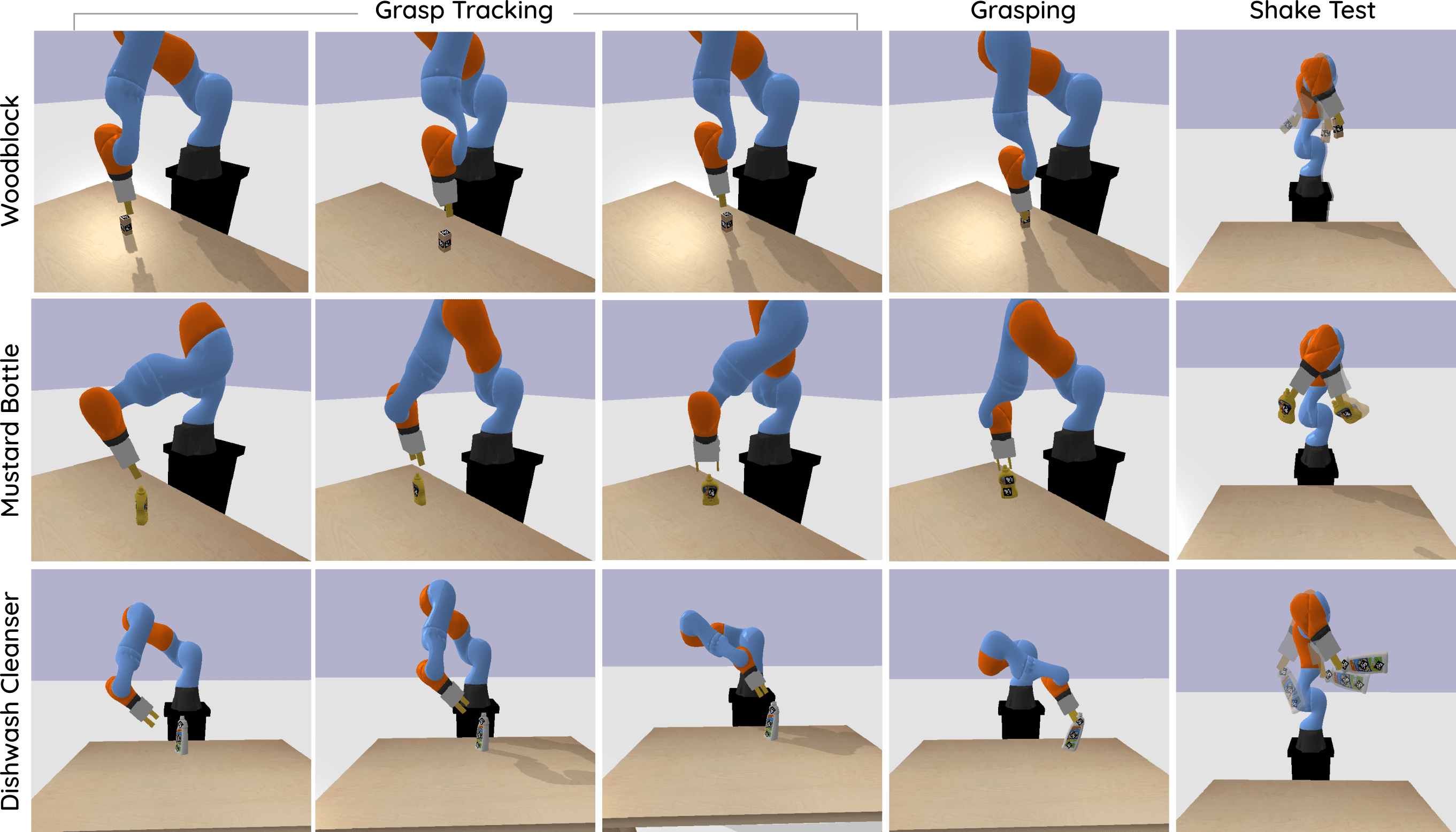}
    \caption{Screenshots depicting the robotic grasping of moving objects. The first three columns show various positions during the trajectory and the fourth column images show object being grasped and the final column images show a grasp stability test.}
    \label{fig:all_grasps}
    \myfigureshrinker
\end{figure*}
The second set of experiments are conducted to evaluate the performance of our method in grasping arbitrarily moving objects. For this, we have identified 5 different trajectories for the object to move: (i) horizontal line from left side of the table to the right; (ii) vertical line, \textit{i.e.}, object moving away from the robot; (iii) diagonal line, from left bottom to right upper corner; (iv) ellipsoid with radius $(15, 8)~\mathrm{cm}$; (v) sine wave with an amplitude and frequency of $15~\mathrm{cm}$ and $0.02~\mathrm{Hz}$, respectively. For each trajectory, we do two different tests: (i) we move the object in the space without any rotations; and (ii) we move the object with added arbitrary rotations.

Our visual servoing is initialised when the LoCoMo provides initial best grasp and pre-grasp poses. By using the pre-grasp pose as a reference pose, the robot starts moving towards it. This is when the object is moved in one of the five aforementioned trajectories. While the object moves, we constantly monitor the error between the desired pre-grasp configuration and the current end-effector pose. Once the sum of the squared error reaches a threshold, we update the desired pose to the best grasp pose. When the controller converges, the robot will execute the grasp with a force of $50~\mathrm{N}$. As mentioned, this two-point approach reduces the chances that the robot collides with objects. Once grasped, the object movement is stopped, and the grasp is evaluated for stability. For this, we conduct the stability tests defined in \cite{BenchmarkERL}. They are: (i) lifting test, in which we lift the object $20~\mathrm{cm}$ above the table at the speed of $10~\mathrm{cm/s}$; (ii) rotation test, where we move the manipulator to be in $90-90$ configuration \cite{BenchmarkERL} and rotate the object from $90^{\circ}$ to $-90^{\circ}$ with a speed of $45~\mathrm{deg/s}$; and finally, (iii) shake test, where we shake the object in a sinusoidal pattern with an amplitude of $0.25~\mathrm{m}$ and a peak acceleration of $10~ \mathrm{m/s^2}$. These three tests are performed sequentially as in the order mentioned, and if the object slips from the gripper at any point, they are deemed failure and the next test is not performed.

Fig. \ref{fig:all_grasps} shows some screenshots of grasping moving objects using our method. Obtained results for all three objects, \textit{i.e.}, woodblock (YCB id: 45), mustard bottle (YCB id: 9), and scrub cleanser (YCB id: 20), are summarised in the Tables \ref{tab:grasping_woodblock}, \ref{tab:grasping_cleanser} and \ref{tab:grasping_mustard}, respectively. We report the number of times the controller has switched the grasping configuration (re-ranking), the time from the start of the control loop until grasping, and the grasp success evaluation. Furthermore, we present the results with and without the null space control activated as well as the results with and without grasp re-ranking enabled. % For each trajectory, we repeat the test three times and the average of these three trials are reported in tables.
 %
%=========== TABLE 1 ============
\midsepremove
\begin{table*}
\smaller
    % \caption{Grasping for the "woodblock" YCB object with the proposed method for different trajectories with and without null-space and with and without best grasp re-ranking}
    \caption{Dynamic grasping results with the ``Woodblock’’ object for various testing trajectories and for various test cases.}
    \label{tab:grasping_woodblock}
    \centering
    \begin{threeparttable}
    \begin{tabularx}{\textwidth}{>{\hsize=0.14\hsize\raggedright\arraybackslash}X|
                              >{\hsize=0.06\hsize\centering\arraybackslash}X|
                              >{\hsize=0.06\hsize\centering\arraybackslash}X|
                             >{\hsize=0.06\hsize\centering\arraybackslash}X|
                             >{\hsize=0.06\hsize\centering\arraybackslash}X|
                             >{\hsize=0.06\hsize\centering\arraybackslash}X|
                              >{\hsize=0.06\hsize\centering\arraybackslash}X|
                              >{\hsize=0.06\hsize\centering\arraybackslash}X|
                             >{\hsize=0.06\hsize\centering\arraybackslash}X|
                             >{\hsize=0.06\hsize\centering\arraybackslash}X|
                             >{\hsize=0.06\hsize\centering\arraybackslash}X|
                             >{\hsize=0.06\hsize\centering\arraybackslash}X|
                              >{\hsize=0.06\hsize\centering\arraybackslash}X|
                             >{\hsize=0.06\hsize\centering\arraybackslash}X|
                             >{\hsize=0.06\hsize\centering\arraybackslash}X|
                             >{\hsize=0.06\hsize\centering\arraybackslash}X}
\toprule
    \multirow{2}{*}{\textbf{Trajectory}} & \multicolumn{5}{c|}{\textbf{Full Method}} & \multicolumn{5}{c|}{\textbf{Without NullSpace}} & \multicolumn{5}{c}{\textbf{Without Re-rank}} \\
    \cmidrule{2-16}
      & \textbf{Switch\tnote{1}} & \textbf{Time\tnote{2}} & \textbf{Lift\tnote{3}} & \textbf{Rot.\tnote{3}} & \textbf{Shake\tnote{3} } & \textbf{Switch\tnote{1}} & \textbf{Time\tnote{2}} & \textbf{Lift \tnote{3} } & \textbf{Rot.\tnote{3}  } & \textbf{Shake\tnote{3}  } & \textbf{Switch\tnote{1}} & \textbf{Time\tnote{2}} & \textbf{Lift\tnote{3}  } & \textbf{Rot.\tnote{3} } & \textbf{Shake \tnote{3} }\\
     \midrule
     H. Line & 1 & 28.81 &100 &100 &100 & 0 & 11.07 & 100  & 100 & 100 & NA & 26.54 & 100 & 100 & 100\\
     H. Line + Rot. & 1 & 32.63 &100 &100 &100 & 3 & - &- &- &- & NA & 30.42 &100 &100 &100\\
     V. Line & 2 & 19.85 &100 &0 &0 & 1 & 72.42 & 0  & 0 & 0 & NA &10.93 & 0 & 0 & 0 \\
     V. Line + Rot. & 1 & 19.313 &100 &100 &100 & 1 & 29.71 & 0  & 0 & 0 & NA & - & -&- & -\\
     Diag. & 1 &19.63 &100 &100 &100 & 3 & 17.15 &100 &100 &100 & NA & - & - & - & -\\
     Diag. + Rot.& 1 & 22.28  &100 &100 &100 & 0 & 20.82 & 100 & 100 & 100 & NA & 29.4 & 100 & 100 & 100\\
     Circle  & 0 & 88.29 &100 &100 &100 & 0 & 11.32 & 100 & 100 & 100 & NA & 81.11& 100 & 100 & 100\\
     Circle + Rot. & 3 & 91.32 &100 &100 &100 &  4 & 97.1 & 0 & 0 & 0 &NA & 79.27 & 100 & 100&100 \\
     Wave  & 2 & 52.929 & 100 &0 &0 & 9 & 161.77 & 100 &100 & 100 & NA & - &- &- &-\\
     Wave + Rot. & 4 & 77.35 &100 &100 &100 & 7 & 134.11 & 100 & 100 & 100 & NA & - & - & - & -\\

     \bottomrule
\end{tabularx}
    \begin{tablenotes}
        \item[1] Number of times the re-ranking switched the reference pose.
        \item[2] Time in seconds from the start pose until the object is grasped.
        \item[3] Percentage of success $\left(\bm{\%}\right)$ of lifting, Rotational and Shaking tests.
    \end{tablenotes}
    \end{threeparttable}
    \myfigureshrinker
\end{table*}

%=========== TABLE 2 ============
\midsepremove
\begin{table*}
\smaller
    % \caption{Grasping for the "Bleach Cleanser" YCB object with the proposed method for different trajectories with and without null-space and with and without best grasp re-ranking}
        \caption{Dynamic grasping results with the ``Scrub Cleanser ’’ object for various testing trajectories and for various test cases.}
    \label{tab:grasping_cleanser}
    \centering
    \begin{threeparttable}
    \begin{tabularx}{\textwidth}{>{\hsize=0.14\hsize\raggedright\arraybackslash}X|
                              >{\hsize=0.06\hsize\centering\arraybackslash}X|
                              >{\hsize=0.06\hsize\centering\arraybackslash}X|
                             >{\hsize=0.06\hsize\centering\arraybackslash}X|
                             >{\hsize=0.06\hsize\centering\arraybackslash}X|
                             >{\hsize=0.06\hsize\centering\arraybackslash}X|
                              >{\hsize=0.06\hsize\centering\arraybackslash}X|
                              >{\hsize=0.06\hsize\centering\arraybackslash}X|
                             >{\hsize=0.06\hsize\centering\arraybackslash}X|
                             >{\hsize=0.06\hsize\centering\arraybackslash}X|
                             >{\hsize=0.06\hsize\centering\arraybackslash}X|
                             >{\hsize=0.06\hsize\centering\arraybackslash}X|
                              >{\hsize=0.06\hsize\centering\arraybackslash}X|
                             >{\hsize=0.06\hsize\centering\arraybackslash}X|
                             >{\hsize=0.06\hsize\centering\arraybackslash}X|
                             >{\hsize=0.06\hsize\centering\arraybackslash}X}
\toprule
    \multirow{2}{*}{\textbf{Trajectory}} & \multicolumn{5}{c|}{\textbf{Full Method}} & \multicolumn{5}{c|}{\textbf{Without NullSpace}} & \multicolumn{5}{c}{\textbf{Without Re-rank}} \\
    \cmidrule{2-16}
      & \textbf{Switch } & \textbf{Time } & \textbf{Lift } & \textbf{Rot. } & \textbf{Shake  } & \textbf{Switch } & \textbf{Time } & \textbf{Lift   } & \textbf{Rot.     } & \textbf{Shake     } & \textbf{Switch} & \textbf{Time} & \textbf{Lift     } & \textbf{Rot.    } & \textbf{Shake     }\\
     \midrule
     H. Line & 2 & 30.85 &100 &100 &100 & 3 & 48.15 & 100  & 100 & 100 & NA & 48.8 & 100 & 100 & 100\\
     H. Line + Rot. & 2 & 39.92 &100 &100 &100 & 3 & 38.39 & 100 & 100 & 100 & NA & - & - & - & -\\
     V. Line & 2 & 43.66 & 100 &0 &0 & 2 & 49.87 & 0  & 0 & 0 & NA & - & - & - & - \\
     V. Line + Rot. & 2 & 37.76 & 100 &100 & 0 & 2 & 31,08 & 100  & 100 & 0 & NA & - & -&- & -\\
     Diag. & 4 & 110.26 &100 &100 &100 & 3 & 39.55 &100 &100 &100 & NA & - & - & - & -\\
     Diag. + Rot.& 2 & 31.45  &100 &100 &100 & 2 & 50.41 & 100 & 100 & 100 & NA & - & 100 & - & -\\
     Circle  & 3 & 76.81 &100 &100 &100 & 0 & 62.44 & 0 & 0 & 0 & NA & 81.11& 100 & 100 & 100\\
     Circle + Rot. & 5 & 88.07 &100 &100 &100 &  4 & 214.04 & 100 & 0 & 0 &NA & 82.438 & 100 & 100 & 100 \\
     Wave  & 0 & 82.438 & 100 & 100 & 100 & 2 & 46.29 & 100 &100 & 100 & NA & 41.29 & 100 & 100 & 100\\
     Wave + Rot. & 2 & 35.73 &100 &100 &100 & 2 & 56.1 & 100 & 100 & 100 & NA & 156.32 & 0 & 0 & 0\\
     \bottomrule
\end{tabularx}
    \end{threeparttable}
   % \mytableshrinker
\end{table*}

%=========== TABLE 3 ============
\midsepremove
\begin{table*}
\smaller
    % \caption{Grasping for the "mustard bottle" YCB object with the proposed method for different trajectories with and without null-space and with and without best grasp re-ranking}
    \caption{Dynamic grasping results with the ``Mustard Bottle’’ object for various testing trajectories and for various test cases.}
    \label{tab:grasping_mustard}
    \centering
    \begin{threeparttable}
    \begin{tabularx}{\textwidth}{>{\hsize=0.14\hsize\raggedright\arraybackslash}X|
                              >{\hsize=0.06\hsize\centering\arraybackslash}X|
                              >{\hsize=0.06\hsize\centering\arraybackslash}X|
                             >{\hsize=0.06\hsize\centering\arraybackslash}X|
                             >{\hsize=0.06\hsize\centering\arraybackslash}X|
                             >{\hsize=0.06\hsize\centering\arraybackslash}X|
                              >{\hsize=0.06\hsize\centering\arraybackslash}X|
                              >{\hsize=0.06\hsize\centering\arraybackslash}X|
                             >{\hsize=0.06\hsize\centering\arraybackslash}X|
                             >{\hsize=0.06\hsize\centering\arraybackslash}X|
                             >{\hsize=0.06\hsize\centering\arraybackslash}X|
                             >{\hsize=0.06\hsize\centering\arraybackslash}X|
                              >{\hsize=0.06\hsize\centering\arraybackslash}X|
                             >{\hsize=0.06\hsize\centering\arraybackslash}X|
                             >{\hsize=0.06\hsize\centering\arraybackslash}X|
                             >{\hsize=0.06\hsize\centering\arraybackslash}X}
\toprule
    \multirow{2}{*}{\textbf{Trajectory}} & \multicolumn{5}{c|}{\textbf{Full Method}} & \multicolumn{5}{c|}{\textbf{Without NullSpace}} & \multicolumn{5}{c}{\textbf{Without Re-rank}} \\
    \cmidrule{2-16}
      & \textbf{Switch} & \textbf{Time   } & \textbf{Lift   } & \textbf{Rot.   } & \textbf{Shake    } & \textbf{Switch} & \textbf{Time   } & \textbf{Lift     } & \textbf{Rot.     } & \textbf{Shake     } & \textbf{Switch} & \textbf{Time   } & \textbf{Lift     } & \textbf{Rot.    } & \textbf{Shake     }\\
     \midrule
     H. Line & 0 & 26.681 &100 &0 & 0 & 2 & 59.09 & 100  & 100 & 100 & NA & 49.9 & 100 & 100 & 0\\
     H. Line + Rot. & 0 & 40.94 &100 &100 &100 & 3 & 38.39 & 100 & 100 &100 & NA & 37.88 &100 &100 &0\\
     V. Line & - & - & - & - & - & - & - & -  & - & - & NA & 52.32 & 0 & 0 & 0 \\
     V. Line + Rot. & 2 & 91.75 &100 &100 &100 & 2 & 97.56 & 100  & 100 & 100 & NA & 65.49 & 100 & 100 & 100 \\
     Diag. & 2 &45.31 &100 &100 &100 & 5 & 83.55 &100 &0 &0 & NA & - & - & - & -\\
     Diag. + Rot.& 1 & 45.31  &100 &100 &100 & 1 & 32.90 & 100 & 0 & 0 & NA & - & - & - & -\\
     Circle  & 5 & 88.07 &100 &100 &100 & 7 & 197.1 & 0 & 0 & 0 & NA & 145.87& 100 & 100 & 100\\
     Circle + Rot. & 3 & 179.42 &100 &100 &100 &  3 & 66.4 & 100 & 100 & 100 &NA & - & - & - & - \\
     Wave  & 1 & 29.83 & 100 &100 &0 & 1 & 47.8 & 0 & 0 & 0 & NA & 48.09 & 100 & 100 & 0\\
     Wave + Rot. & 3 &45.69 &100 &100 &100 & 3 & 0 & 0 & 0 & 0 & NA & 48.57 & 100 & 100 & 100 \\
     \bottomrule
\end{tabularx}
    \end{threeparttable}
    \mytableshrinker
\end{table*}

Undoubtedly, the full method with null space control activated and re-ranking enabled, outperformed the others. On average, for all three objects respectively, the full method achieved $[\bm{80\%}, \bm{80\%}, \bm{70\%}]$ success while the method without null space achieved $[60\%, 60\%, 40\%]$ success and the method without re-ranking enabled achieved $[50\%, 30\%, 30\%]$ success. With full method, for the first two objects we got $\bm{100\%}$ success rate following and grasping the objects. We note that, due to joint constraints, the controller without the null space did not converge to grasp the object. We also note that the number of switches is higher without the null space. That is because when the null space control is deactivated, the re-ranking compensates unfeasible configurations. Out of all cases, the method without re-ranking showcased poor performance. We believe that this is due to the lack of inverse kinematics during trajectories. Overall, the obtained results clearly demonstrate the efficiency of our proposed dual quaternion-based visual servoing in grasping objects that are moving arbitrarily in a 3D space.

%% file: sections/conclusion.tex
In this paper, we have presented a method for grasping moving objects by formulating a visual servoing controller based on dual quaternion algebra. For the problem of grasping we have  incorporated a novel re-ranking strategy to the LoCoMo grasp planner, so that grasp candidates automatically update whenever the object moves. The dual quaternion proves to be advantageous when performing visual servoing for grasping. The coupled rotation and translation allow the manipulator to quickly converge to a desired grasp pose even in the presence of perturbations. %Furthermore, the problem of grasping moving objects can result in the system running close to singular configurations or joint limits. 
Through simulation experiments, we have demonstrated that in combination with joint limit avoidance and dynamic re-ranking, our method demonstrates high success rate in grasping moving objects. In future works we plan on improving the control strategy with optimisation techniques. Furthermore, we also plan on extending this work by performing real robot experiments and integrating the object tracking with dual quaternions.%tackling the object tracking problem without the use of markers.

%% file: main.bbl
\begin{thebibliography}{10}
\providecommand{\url}[1]{#1}
\csname url@samestyle\endcsname
\providecommand{\newblock}{\relax}
\providecommand{\bibinfo}[2]{#2}
\providecommand{\BIBentrySTDinterwordspacing}{\spaceskip=0pt\relax}
\providecommand{\BIBentryALTinterwordstretchfactor}{4}
\providecommand{\BIBentryALTinterwordspacing}{\spaceskip=\fontdimen2\font plus
\BIBentryALTinterwordstretchfactor\fontdimen3\font minus
  \fontdimen4\font\relax}
\providecommand{\BIBforeignlanguage}[2]{{%
\expandafter\ifx\csname l@#1\endcsname\relax
\typeout{** WARNING: IEEEtran.bst: No hyphenation pattern has been}%
\typeout{** loaded for the language `#1'. Using the pattern for}%
\typeout{** the default language instead.}%
\else
\language=\csname l@#1\endcsname
\fi
#2}}
\providecommand{\BIBdecl}{\relax}
\BIBdecl

\bibitem{bicchi2000robotic}
A.~{Bicchi} and V.~{Kumar}, ``Robotic grasping and contact: a review,'' in
  \emph{IEEE Int. Conf. on Rob. and Auto.}, vol.~1, 2000, pp. 348--353 vol.1.

\bibitem{sahbani2012overview}
A.~Sahbani \emph{et~al.}, ``An overview of 3d object grasp synthesis
  algorithms,'' \emph{Rob. Auton. Syst.}, vol.~60, no.~3, pp. 326--336, 2012.

\bibitem{bohg2013data}
J.~Bohg \emph{et~al.}, ``Data-driven grasp synthesis—a survey,'' \emph{{IEEE}
  Trans. Robot.}, vol.~30, no.~2, pp. 289--309, 2013.

\bibitem{Farias_BayesianGrasping}
C.~de~Farias \emph{et~al.}, ``Simultaneous tactile exploration and grasp
  refinement for unknown objects,'' \emph{IEEE Robotics and Automation
  Letters}, vol.~6, no.~2, pp. 3349--3356, 2021.

\bibitem{2006_chaumette_visual_P1}
F.~Chaumette and S.~Hutchinson, ``Visual servo control. i. basic approaches,''
  \emph{IEEE Robot. Autom. Mag}, vol.~13, pp. 82--90, 2006.

\bibitem{marchand2015pose}
E.~Marchand \emph{et~al.}, ``Pose estimation for augmented reality: a hands-on
  survey,'' \emph{IEEE Trans. Vis. Comput. Graphics}, vol.~22, no.~12, pp.
  2633--2651, 2015.

\bibitem{marturi2019dynamic}
N.~Marturi \emph{et~al.}, ``Dynamic grasp and trajectory planning for moving
  objects,'' \emph{Autonomous Robots}, vol.~43, no.~5, pp. 1241--1256, 2019.

\bibitem{gridseth2015visual}
M.~Gridseth \emph{et~al.}, ``On visual servoing to improve performance of
  robotic grasping,'' in \emph{12th Conference on Computer and Robot
  Vision}.\hskip 1em plus 0.5em minus 0.4em\relax IEEE, 2015, pp. 245--252.

\bibitem{husain2014realtime}
F.~Husain \emph{et~al.}, ``Realtime tracking and grasping of a moving object
  from range video,'' in \emph{Proc. IEEE Int. Conf. Robot. Autom.}\hskip 1em
  plus 0.5em minus 0.4em\relax IEEE, 2014, pp. 2617--2622.

\bibitem{kim2014catching}
S.~Kim \emph{et~al.}, ``Catching objects in flight,'' \emph{IEEE Trans.
  Robot.}, vol.~30, no.~5, pp. 1049--1065, 2014.

\bibitem{1999_Daniilidis_HandEyeCalib}
K.~Daniilidis, ``Hand-eye calibration using dual quaternions,'' \emph{Int. J.
  Rob. Res.h}, vol.~18, no.~3, pp. 286--298, 1999.

\bibitem{2018_LI_Slam}
K.~{Li} \emph{et~al.}, ``Simultaneous localization and mapping using a novel
  dual quaternion particle filter,'' in \emph{21st Int. Conf. on Information
  Fusion}, 2018, pp. 1668--1675.

\bibitem{2020_Saltus_DQ_PVBS}
R.~{Saltus} \emph{et~al.}, ``Dual quaternion visual servo control,'' in
  \emph{Proc. IEEE Conf. Decis. Control.}, 2020, pp. 5956--5961.

\bibitem{2005Wu_navigation}
{Yuanxin Wu} \emph{et~al.}, ``Strapdown inertial navigation system algorithms
  based on dual quaternions,'' \emph{IEEE Trans. Aerosp. Electron. Syst.},
  vol.~41, no.~1, pp. 110--132, 2005.

\bibitem{busam2017camera}
B.~Busam \emph{et~al.}, ``Camera pose filtering with local regression geodesics
  on the riemannian manifold of dual quaternions,'' in \emph{Proc. IEEE Int.
  Conf. Comp. Vis. Workshops}, 2017, pp. 2436--2445.

\bibitem{2018_Adjigble_LoCoMo}
M.~{Adjigble} \emph{et~al.}, ``Model-free and learning-free grasping by local
  contact moment matching,'' in \emph{IEEE/RSJ International Conference on
  Intelligent Robots and Systems}, 2018, pp. 2933--2940.

\bibitem{garrido2014automatic}
S.~Garrido-Jurado \emph{et~al.}, ``Automatic generation and detection of highly
  reliable fiducial markers under occlusion,'' \emph{Pattern Recognition},
  vol.~47, no.~6, pp. 2280--2292, 2014.

\bibitem{BenchmarkERL}
Y.~{Bekiroglu} \emph{et~al.}, ``Benchmarking protocol for grasp planning
  algorithms,'' \emph{{IEEE} Robot. Autom. Lett.}, vol.~5, no.~2, pp. 315--322,
  2020.

\bibitem{hamilton1866elements}
W.~R. Hamilton, \emph{Elements of quaternions}.\hskip 1em plus 0.5em minus
  0.4em\relax Longmans, Green, \& Company, 1866.

\bibitem{selig2004geometric}
J.~M. Selig, \emph{Geometric fundamentals of robotics}.\hskip 1em plus 0.5em
  minus 0.4em\relax Springer Science \& Business Media, 2004.

\bibitem{Chaumette07b}
F.~Chaumette, ``Visual servoing,'' in \emph{Robot Manipulators: Modeling,
  Performance Analysis and Control}, E.~Dombre and W.~Khalil, Eds.\hskip 1em
  plus 0.5em minus 0.4em\relax ISTE, 2007, ch.~6, pp. 279--336.

\bibitem{han2008_quaternionControl}
D.~{Han} \emph{et~al.}, ``A dual-quaternion method for control of spatial rigid
  body,'' in \emph{IEEE Int. Conf. on Networking, Sensing and Control}, 2008,
  pp. 1--6.

\bibitem{2018Figueredo}
L.~F.~C. Figueredo \emph{et~al.}, ``{Robust H-infinity kinematic control of
  manipulator robots using dual quaternion algebra},'' \emph{Automatica}, pp.
  1--8, Jun. 2021.

\bibitem{adjigble2019assisted}
M.~{Adjigble} \emph{et~al.}, ``An assisted telemanipulation approach: combining
  autonomous grasp planning with haptic cues,'' in \emph{IEEE/RSJ Int. Conf.
  Intell. Rob. Sys.}, 2019, pp. 3164--3171.

\bibitem{yianilos1993data}
P.~N. Yianilos, ``Data structures and algorithms for nearest neighbor search in
  general metric spaces,'' in \emph{Proc. Fourth annual ACM-SIAM Symposium on
  Discrete algorithms}, 1993, pp. 311--321.

\bibitem{coumans2016pybullet}
E.~Coumans and Y.~Bai, ``Pybullet, a python module for physics simulation for
  games, robotics and machine learning,'' 2016.

\bibitem{ycb}
B.~Calli \emph{et~al.}, ``The ycb object and model set: Towards common
  benchmarks for manipulation research,'' in \emph{Proc. Int. Conf. Adv.
  Robot.}\hskip 1em plus 0.5em minus 0.4em\relax IEEE, 2015, pp. 510--517.

\end{thebibliography}
